\documentclass[10pt,twocolumn,letterpaper]{article}

\usepackage{iccv}
\usepackage{times}
\usepackage{epsfig} 
\usepackage{graphicx}
\usepackage{amsmath}
\usepackage{amssymb}

\usepackage{diagbox}
\usepackage{subfigure} 
\usepackage{authblk}

\usepackage[pagebackref=true,breaklinks=true,letterpaper=true,colorlinks,bookmarks=false]{hyperref}

\iccvfinalcopy 

\def\iccvPaperID{2526} 

\ificcvfinal\fi
\begin{document}

\title{LFFD: A Light and Fast Face Detector for Edge Devices}
\author[1,2]{Yonghao He\footnote[1]}
\author[1]{Dezhong Xu\footnote[1]} 
\author[1]{Lifang Wu}
\author[1]{Meng Jian}
\author[2]{Shiming Xiang}
\author[2]{Chunhong Pan}
\affil[1]{Faculty of Information Technology, Beijing University of Technology}
\affil[2]{National Laboratory of Pattern Recognition, Institute of Automation, Chinese Academy of Sciences\authorcr 
\tt{\small{yonghao.he@aliyun.com,xudezhong@emails.bjut.edu.cn,lfwu@bjut.edu.cn,}\authorcr 
\tt{jianmeng648@163.com,\{smxiang, chpan\}@nlpr.ia.ac.cn}}}

\maketitle

\begin{abstract}
Face detection, as a fundamental technology for various applications, is always deployed on edge devices which have limited memory storage and low computing power. 
This paper introduces a Light and Fast Face Detector (\textbf{LFFD}) for edge devices.
The proposed method is anchor-free and belongs to the one-stage category.
Specifically, we rethink the importance of receptive field (RF) and effective receptive field (ERF) in the background of face detection.
Essentially, the RFs of neurons in a certain layer are distributed regularly in the input image and theses RFs are natural ``anchors".
Combining RF ``anchors" and appropriate RF strides, the proposed method can detect a large range of continuous face scales with 100\% coverage in theory.
The insightful understanding of relations between ERF and face scales motivates an efficient backbone for one-stage detection.
The backbone is characterized by eight detection branches and common layers, resulting in efficient computation.
Comprehensive and extensive experiments on popular benchmarks: WIDER FACE and FDDB are conducted. 
A new evaluation schema is proposed for application-oriented scenarios.
Under the new schema, the proposed method can achieve superior accuracy (WIDER FACE Val/Test -- Easy: 0.910/0.896, Medium: 0.881/0.865, Hard: 0.780/0.770; FDDB -- discontinuous: 0.973, continuous: 0.724). 
Multiple hardware platforms are introduced to evaluate the running efficiency. 
The proposed method can obtain fast inference speed (
NVIDIA TITAN Xp: 131.45 FPS at 640$\times$480;
NVIDIA TX2: 136.99 PFS at 160$\times$120;
Raspberry Pi 3 Model B+: 8.44 FPS at 160$\times$120) with model size of 9 MB. 
\end{abstract}
\renewcommand{\thefootnote}{\fnsymbol{footnote}}
\footnotetext[1]{Authors contributed equally.}

\section{Introduction}
\label{introduction}
Face detection is a long-standing problem in computer vision. 
In practice, it is the prerequisite to some face-related applications, such as face alignment~\cite{FAS} and face recognition~\cite{DFRS}.
Besides, face detectors are always deployed on edge devices, such as mobile phones, IP cameras and IoT (Internet of Things) sensors. 
These devices have limited memory storage and low computing power.
Under such condition, face detectors that have high accuracy and fast running speed are in demand.

\begin{table}
\label{table-top5-widerface}
\begin{center}
\begin{tabular}{|c|c|c|c|}
\hline
\diagbox{\textbf{Method}}{\textbf{mAP(\%)}}{\textbf{Subset}} & Easy & Medium & Hard \\
\hline
ISRN\cite{ISRN} 						& 0.967 & 0.958 & 0.909\\
VIM-FD\cite{VIM-FD} 				& 0.967 & 0.957 & 0.907\\
DSFD\cite{DSFD} 					& 0.966 & 0.957 & 0.904\\
SRN\cite{SRN} 						& 0.964 & 0.952 & 0.901\\
PyramidBox\cite{PyramidBox} & 0.961 & 0.950 & 0.889\\
\hline
\end{tabular}
\end{center}
\caption{Accuracy of the top-5 methods on validation set of WIDER FACE.}
\end{table}
Current state of the art face detectors have achieved fairly high accuracy on convictive benchmark WIDER FACE~\cite{WIDER} by leveraging pre-trained heavy backbones like VGG16~\cite{VGG}, Resnet50/152~\cite{Resnet} and Densenet121~\cite{Densenet}. 
We investigate the top-5 methods on WIDER FACE and present their accuracy in Table~\ref{table-top5-widerface}. 
It can be observed that these methods have similar accuracy with marginal gaps which are hardly perceived in practical applications.
It is difficult and unpractical to further boost the accuracy by using more complex and heavier backbones. 
In our view, to better balance accuracy and latency is crucial for applying face detection to more applicable areas.

Face detection is a fast-growing branch of general object detection in the past decade. 
The early work of Viola-Jones face detector~\cite{Viola-Jones} proposes a classic detection framework -- cascade classifiers with hand-crafted features. 
One of its well-known followers is aggregate channel features (ACF) \cite{ACF,ACF-Face} which can take advantages of channel features effectively. 
Although the methods mentioned above can achieve fast running speed, they rely on hand-crafted features and are not trained end-to-end, resulting in not robust detection accuracy.

Recently, convolutional neural network (CNN) based face detectors~\cite{ISRN,VIM-FD,DSFD,SRN,PyramidBox,FFRCNN,FRCNN,MTCNN,HR,S3FD,SSF,SSH,FaceBoxes} show great progress partially owing to the success of WIDER FACE benchmark.
These methods can be roughly divided into two categories: two-stage methods and one-stage methods.
Two-stage methods~\cite{FFRCNN,FRCNN} consist of proposal selection and localization regression, which are mainly originated from R-CNN series~\cite{RCNN,FastRCNN,FasterRCNN}.
Whereas, one-stage methods~\cite{HR,S3FD,SSH,FaceBoxes,PyramidBox,SRN,DSFD,ISRN} coherently combine classification and bounding box (bbox) regression, always achieving anchor-based and multi-scale detection simultaneously.
For most one-stage methods, anchor design and matching strategy is one of the essential components.
In order to improve the accuracy, these methods propose more complex modules based on heavy backbones.
Although the above methods can achieve state of the art results, they may not properly balance accuracy and latency.

In this paper, we propose a Light and Fast Face Detector (\textbf{LFFD}) for edge devices, considerably balancing both accuracy and running efficiency.
The proposed method is inspired by the one-stage and multi-scale object detection method SSD~\cite{SSD} which also enlightens some other face detectors~\cite{DSFD,PyramidBox,S3FD}.
One of the characteristics of SSD is that pre-defined anchor boxes are manually designed for each detection branch.
These boxes always have different sizes and aspect ratios to cover objects with different scales and shapes.
Therefore, anchors play an important role in most one-stage detection methods.
For some face detectors~\cite{S3FD,SSF,PyramidBox,DSFD}, sophisticated anchor strategies are crucial parts of the contributions.
However, anchor based methods may face three challenges: 1) anchor matching is unable to sufficiently cover all face scales. Although this can be relieved, it remains a problem;
2) matching anchors to groundtruth bboxes is determined by thresholding IOU (Intersection over Union). The threshold is set empirically and it is difficult to make a solid investigation of its impact;
3) setting the number of anchors for different scales depends on experiences, which may induce sample imbalance and redundant computation.

In our point of view, RF of neurons in feature maps are inherent and natural ``anchors". RF can easily handle above challenges.
Firstly, continuous scales of faces can be predicted within a certain RF size, rather than discrete scales in anchor-based methods.
Secondly, matching strategy is clear, namely a RF is matched to a groundtruth bbox if and only if its center falls in the groundtruth bbox .
Thirdly, the number of RFs is naturally fixed and they are regularly distributed in the input image.
What's more, we make a qualitative analysis on pairing face scales and RF sizes by understanding the insights of ERF,
resulting in an efficient backbone with eight detection branches. 
The backbone only consists of common layers (conv3$\times$3, conv1$\times$1, ReLU and residual connection), which is much lighter than VGG16~\cite{VGG}, Resnet50~\cite{Resnet} and Densenet121~\cite{Densenet}.
Consequently, the final model has only 2.1M parameters ( versus VGG16-138.3M and Resnet50-25.5M ) and achieves superior accuracy and running speed, which makes it appropriate for edge devices.

In summary, the main contributions of this paper include:
\begin{itemize}
\setlength{\itemsep}{0pt}
\setlength{\parsep}{0pt}
\setlength{\parskip}{0pt}
\item We study the relations of RF, ERF and face detection. The relevant understanding motivates the network design.
\item We introduce the RF to overcome the drawbacks of the previous anchor-based strategies, resulting in a anchor-free method.
\item We proposed a new backbone with common layers for accurate and fast face detection.
\item Extensive and comprehensive experiments on multiple hardware platforms are conducted on benchmarks WIDER FACE and FDDB to firmly demonstrate the superiority of the proposed method for edge devices.
\end{itemize}

\section{Related Work}
\label{related_work}
Face detection has attracted a lot of attention since a decade ago. 

\textbf{Early works}
~Early face detectors leverage hand-crafted features and cascade classifiers to detect faces in forms of sliding window.
Viola-Jones face detector~\cite{Viola-Jones} uses Adaboost with Haar-like features to train face classifiers discriminatively.
Subsequently, utilizing more effective hand-crafted features~\cite{LBP,HOG,ACF-Face} and more powerful classifiers~\cite{CBE,SBFD} becomes the mainstream.
These methods are not trained end-to-end, treating feature learning and classifier training separately.
Although achieving fast running speed, they can not obtain satisfied accuracy.

\textbf{CNN-based methods}
~Current CNN-based face detectors benefit from two-stage~\cite{RCNN,FastRCNN,FasterRCNN} and one-stage~\cite{SSD,YOLO,YOLOv2,YOLOv3} general object detection.
Both~\cite{FFRCNN} and~\cite{FRCNN} are based on faster R-CNN~\cite{FasterRCNN}, adapting the original faster R-CNN to face detection.
Zhang \etal \cite{ICS} proposes a cascaded CNN for coarse-to-fine face detection with inside cascaded structure.
Recently, one-stage face detectors are dominant.
MTCNN~\cite{MTCNN} performs face detection in a sliding window manner and relies on image pyramid.
HR~\cite{HR} is an advanced version of MTCNN to some extent, also requiring image pyramid.
Image pyramid has some drawbacks like slow speed and high memory cost.
S3FD~\cite{S3FD} takes RF into consideration for detection branch design and proposes an anchor matching strategy to improve hit rate.
In~\cite{SSF}, Zhu \etal focuses on detecting small faces by proposing a robust anchor generating and matching strategy.
It can be concluded that anchor related strategies are crucial for face detection.
Following S3FD~\cite{S3FD}, PyramidBox~\cite{PyramidBox} enhances the backbone with low-level feature pyramid layers (LFPN) for better multi-scale detection.
SSH~\cite{SSH} constructs three detection modules cooperating with context modules for scale-invariant face detection.
DSFD~\cite{DSFD} is characterized by feature enhance modules, early layer supervision and an improved anchor matching strategy for better initialization.
S3FD, PyramidBox, SSH and DSFD use VGG16 as backbones, leading to big model size and inefficient computation.
FaceBoxes~\cite{FaceBoxes} aims to make the face detector run in real-time by rapidly reducing the size of input images.
In detail, it reaches a large stride size 32 after four layers: two convolution layers and two pooling layers.
Although the running speed of FaceBoxes is fast, it abandons the detection of small faces, resulting in relatively low accuracy on WIDER FACE.
Different from FaceBoxes, our method handles the detection of small faces delicately, achieving fast running speed and large scale coverage in the meantime.
It can be observed that the networks used by recent state of the art methods tend to become more complex and heavier.
In our view, to gain marginal improvement in accuracy at the cost of running speed is not appropriate for practical applications.

\section{Light and Fast Face Detector}
\label{light_and_fast_face_detector}
In this section, we first revisit the concept of RF and its relation to face detection in Sec.~\ref{revisit_rf_in_context_of_face_detection}. 
Then Sec.~\ref{rfs_as_natural_anchors} describes the rationality and advantages of using RFs as natural ``anchors".
Subsequently, the details of the proposed network is depicted in Sec.~\ref{backbone_framework}.
Finally, we present the specifications of network training in Sec.~\ref{training_details}.
\subsection{Revisit RF in the Background of Face Detection}
\label{revisit_rf_in_context_of_face_detection}
In the beginning, we make a brief description of RF and its properties.
RF is a definite area of the input image, which affects the activation of the corresponding neuron.
RF determines the range that a neuron can see in the original input.
Intuitively, the target object can be well detected with high probabilities if it is enclosed by a certain RF.
In general, the neurons in shallow layers have small RFs and those in deeper layers have large RFs.
One of the important properties of RF is that each input pixel contributes differently for the neuron's activation~\cite{UERF}.
Specifically, the pixels locating around the center of RF have larger impact. And the impact decreases gradually when the pixels are far away from the center.
This phenomenon is named as effective receptive field (ERF).
ERFs inherently exist in neural networks and present a Gaussian-like distribution.
Thus, making the target object in the middle of the RF is also important. 
The proposed LFFD benefits from the above observations.

\begin{figure}[ht]
\begin{center}
\centerline{\includegraphics[width=7cm]{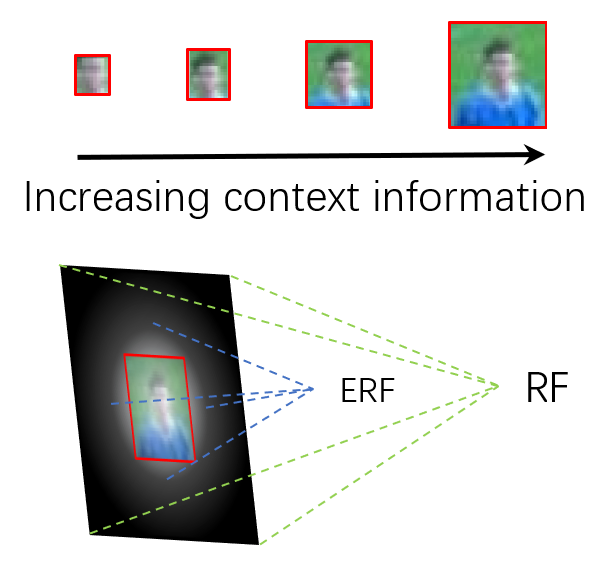}}
\end{center}
\caption{Tiny faces detection. The top-left image only contains a face, and the top-right image depicts a face with sufficient context information.
It is easy to see that the face becomes more distinguishable with the context information gradually increasing.
The lower part describes the relation between RF and ERF for detecting the tiny face. }
\label{fig:RF_and_Face}
\end{figure}

Face detection is a well-known branch of general object detection and it has some characteristics.
First, big faces are approximately rigid due to their unmovable components, such as eyes, noses and mouths. 
Although there are facial expression changes, hair occlusion and other unconstrained situations, big faces are still distinguishable.
Second, tiny or small faces have to be treated differently compared to big faces.
Tiny faces always have unrecognizable appearances (an example is shown in Fig.~\ref{fig:RF_and_Face}).
It is even difficult for humans to make a face/non-face decision by only seeing the facial area of a tiny face, and the same goes for CNN based classifiers\cite{HR}.
With more context information including necks and shoulders, tiny faces become easier to recognize.
Detailed discussion can be referred to \cite{HR}.

Based on above understandings, faces with different sizes need various RF strategies:
\begin{itemize}
\item for tiny/small faces, ERFs have to cover the faces as well as sufficient context information;
\item for medium faces, ERFs only have to contain the faces with little context information;
\item for large faces, only keeping them in RFs is enough.
\end{itemize}
These strategies guide us to design an effective backbone.

\begin{figure*}[!ht]
\begin{center}
\centerline{\includegraphics[width=18cm]{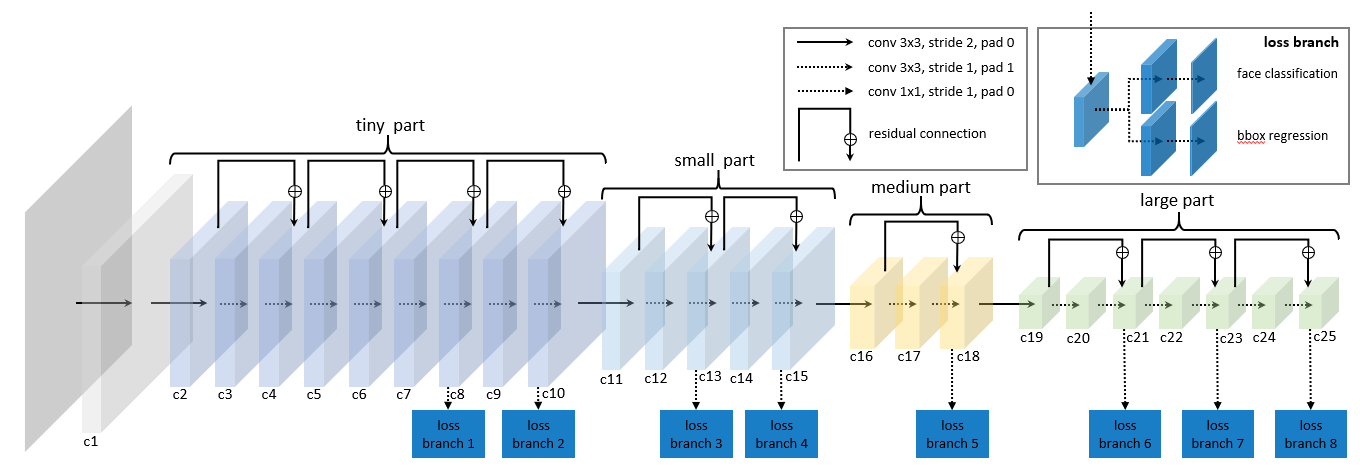}}
\end{center}
\caption{The overall architecture of the proposed network. The backbone has 25 convolution layers and is divided into four parts: tiny part, small part, medium part and large part.
Along the backbone, there are eight loss branches which are in charge of detecting faces with different scales. The entire backbone only consists of conv 3$\times$3, conv 1$\times$1, ReLU and residual connection.}
\label{fig:framwork}
\end{figure*}

\subsection{RFs as Natural ``Anchor''}
\label{rfs_as_natural_anchors}
One-stage detectors are mostly characterized by pre-defined bbox anchors.
In order to detect different objects, anchors are in multiple aspect ratios and sizes.
These anchors are always redundantly defined.
In terms of face detection, it is rational to use 1:1 aspect ratio anchors since faces are approximately square, which is also mentioned in~\cite{S3FD,FaceBoxes}.
The shapes of RFs are also square if the width and height of the kernel are equal. 
The proposed method regards RFs as natural ``anchors".
For the neurons in the same layer, their RFs are regularly tiled in the input image.
The number and size of RFs are inherently determined once the network is built.

As for matching strategy, the proposed method uses a straight and concise way -- the RF is matched to a groundtruth bbox if and only if its center falls in the groundtruth bbox, other than thresholding IOU.
In the typical anchor-based method S3FD~\cite{S3FD}, Zhang \etal also analyses the influence of ERFs and designs anchor augmentation for tiny faces in particular.
In spite of improving the anchor hit rate, S3FD induces the anchor imbalance problem (too many anchors for tiny faces) which has to be addressed by additional means.
However, the proposed method can achieve 100\% face coverage in theory by controlling the RF stride.
Besides, RF with our matching strategy can naturally handle continuous face scales. For an instance, RFs of 100 pixels are able to predict faces between 20 pixels to 40 pixels.
In this way, anchor imbalance problem is greatly relieved and faces from each scale are equally treated.

Based on the above discussion, we do not create any anchors and the proposed method do not really match anchors to groundtruth bboxes.
Therefore, the proposed method is anchor-free.

\begin{table}
\begin{center}
\begin{tabular}{c}
\includegraphics[width=8.2cm]{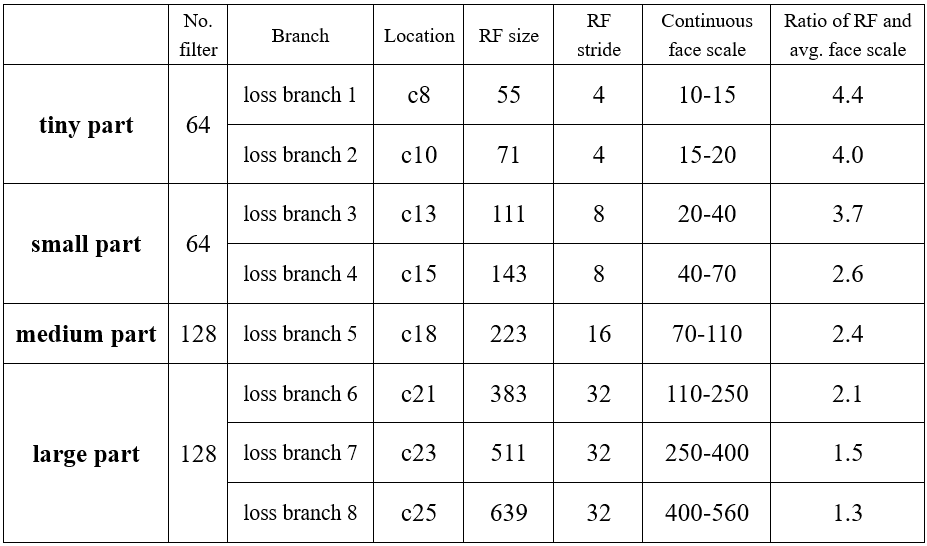}
\end{tabular}
\end{center}
\caption{Detailed information about the proposed network.}
\label{table_details_network}
\end{table}

\subsection{Network Architecture}
\label{backbone_framework}
According to above analyses, we can design a specialised backbone for face detection.
There are two factors that determine the placement of loss branches -- the size and stride of RFs.
The size of RFs guarantees that the learned features of faces are robust and distinguishable,
whereas the stride ensures the 100\% coverage.
The overall architecture of the proposed network is illustrated in Fig.~\ref{fig:framwork}.
The proposed method can detect faces that are lager than 10 pixels (the size of a face is indicated by the longer side), since WIDER FACE benchmark dataset requires faces more than 10 pixels to be detected.
It can be observed that the proposed backbone is one-stage with four parts.
The concrete information about loss branches can be found in Table~\ref{table_details_network}.

The tiny part has 10 convolution layers. 
The first two layers downsample the input with stride 4, stride 2 from each.
Therefore, RFs of other convolution layers in this part are in stride 4.
One crucial principle is: downsample the input as quick as possible while keeping the 100\% face coverage.
This part has two loss branches.
The loss branch 1 stems from c8 whose RF size is 55 for continuous face scale 10-15.
Similarly, the loss branch 2 is from c10 with RF size 71 for continuous face scale 15-20.
Obviously, we can make sure that centers of at least two RFs can fall in the smallest face, thus achieving 100\% coverage.
There is a special case that one center may fall in more than two faces at the same time, in which the corresponding RF is ignored directly.
As we have discussed in Sec.~\ref{revisit_rf_in_context_of_face_detection}, tiny faces need more context information and ERFs are smaller than RFs.
To this end, we use much larger RFs than average face scales. 
The ratios of RFs and average face scales are 4.4 and 4.0 for branch 1 and branch 2, respectively.
In Table~\ref{table_details_network}, such ratios are gradually decreased from 4.4 to 1.3, because larger faces need less context information.
In the backbone, all convolution layers have the kernel size of 3$\times$3. 
Nevertheless, the kernel size of convolution layers in branches is 1$\times$1 which does not change the size of RFs.
In each branch, there are two sub-branches, one for face classification and the other one for bbox regression.

The small part is in charge of two continuous face scales 20-40 and 40-70.
The first convolution layer c11 in this part downsamples the feature maps by 2$\times$.
For the subsequent parts, their first convolution layers accomplish the same function.
In small part, the RF increasing speed becomes 16 compared to that of tiny part 8.
So it takes less convolution layers to reach the targeted RF sizes.
The medium part is similar to the small part, having only one branch.

At the end of the backbone, the large part has seven convolution layers.
These layers easily enlarge the detection scale without too much computation gain due to small feature maps.
Three branches are from this part. 
Since big faces are much easier to detect, the ratios of RFs and average face scales are relatively small.

The proposed method can detect a large range of faces from 10 pixels to 560 pixels within one inference.
The overall backbone only consists of conv 3$\times$3, conv 1$\times$1, ReLU and residual connection.
The main reason is that conv 3$\times$3 and conv 1$\times$1 are highly optimized by inference libraries, such as cuDNN\footnote{https://developer.nvidia.com/cudnn}, ncnn\footnote{https://github.com/Tencent/ncnn}, mace\footnote{https://github.com/XiaoMi/mace} and paddle-mobile\footnote{https://github.com/PaddlePaddle/paddle-mobile}, since they are most widely used.
We do not adopt BN~\cite{BN} as components due to slow inference speed, although it has become the standard configuration of many networks.
We compare the speed between the original backbone and the one with BN: the original one can achieve 7.6 ms and the one with BN only has 8.9 ms, resulting in 17\% slower (resolution: 640$\times$480, hardware: TITAN X (Pascal)) .
In stead of using BN, we train much more iterations for better convergence.
As shown in Fig.~\ref{fig:framwork}, in each part, residual connections are placed side by side for easily training the deep backbone.
The number of filters of all convolution layers in the first two parts is 64.
We do not increase the filters, since the first two parts have relatively large feature maps which are computationally expensive.
However, the number of filters in the last two parts can be increased to 128 without too much additional computation.
More details can be found in Table~\ref{table_details_network}.

\subsection{Training Details}
\label{training_details}
In this subsection, we describe the training related details in several aspects.

\textbf{Dataset and data augmentation.} 
The proposed method is trained on the training set of WIDER FACE benchmark~\cite{WIDER}, including 12,880 images with more than 150,000 valid faces.
Faces less than 10 pixels are discarded directly.
Data augmentation is important for improving the robustness. The detailed strategies are listed as follows:
\begin{itemize}
\setlength{\itemsep}{0pt}
\setlength{\parsep}{0pt}
\setlength{\parskip}{0pt}
 \item Color distort, such as random lighting noise, random contrast, random brightness, \etal. 
 More information can refer to~\cite{SIDCNN,ICDCNN}.
 \item Random sampling for each scale. In the proposed network, there are eight loss branches, each in charge of a certain continuous scale. Thus, we have to guarantee that:
  1) the number of faces for each branch is approximately the same; 2) each face can be sampled for each branch with the same probability.
  To this end, we first randomly select an image, and then randomly select a face in the image. Second, a continuous face scale is selected and the face is randomly resized within the scale as well as the entire image and other
  face bboxes. Finally, we crop a sub-image of 640$\times$640 at the center of the selected face, filling the outer space with black pixels. 
  \item Randomly horizontal flip. We flip the cropped image with probability of 0.5.
 \end{itemize} 

\textbf{Loss function.}
In each loss branch, there are two sub-branches for face classification and bbox regression. 
For face classification, we use softmax with cross-entropy loss over two classes. The matched RF anchors are positive and the others are negative. 
Those RF anchors with more than one matched faces are ignored.
Besides, gray scale is set for each continuous scale. Let $\{SL_i\}_{i=1}^8$ be lower bounds of continuous scales and $\{SU_i\}_{i=1}^8$ for upper bounds. 
The lower and upper gray bounds are calculated as $\{\lfloor SL_i*0.9 \rfloor\}_{i=1}^8$ and $\{\lceil SU_i*1.1 \rceil\}_{i=1}^8$.
For each continuous scale $i$, the relevant gray scales are $[\lfloor SL_i*0.9 \rfloor,SL_i]$ and $[SU_i , \lceil SU_i*1.1 \rceil]$.
For example, branch 3 is for face scale 20-40, the corresponding gray scales are $[18,20]$ and $[40,44]$.
Faces that fall in gray scales are also ignored by the corresponding branch.
For bbox regression, we adopt $L2$ loss directly. The regression groundtruth is defined as:
\begin{eqnarray}
\frac{RF_x - b^{tl}_x}{RF_s/2}, \frac{RF_y - b^{tl}_y}{RF_s/2}, \frac{RF_x - b^{br}_x}{RF_s/2}, \frac{RF_y - b^{br}_y}{RF_s/2}, 
\end{eqnarray}
where $RF_x$ and $RF_y$ are center coordinates of the RF, $b^{tl}_x$ and $b^{tl}_y$ are coordinates of top-left corner of the bbox, $b^{br}_x$ and $b^{br}_y$ are coordinates of bottom-right corner of the bbox and
the normalization constant is $RF_s/2$, $RF_s$ is the RF size.
The $L2$ loss is only activated for positive RF anchors without being ignored.
In the final loss function, the two losses have the same weight.

\textbf{Hard negative mining}.
For each branch, negative RF anchors are usually more than positive ones.
For stable and better training, only a fractional negative RF anchors are used for back-propagation:
we sort the loss values of all negative anchors and only select the top ones for learning.
The ratio between the positive and negative anchors is at most 1:10.
Empirically, hard negative mining can bring faster and stable convergence.

\textbf{Training parameters.}
We initialize all parameters with xavier method and train the network from scratch.
The inputs first minus 127.5, and then divided by 127.5.
The optimization method is SGD with 0.9 momentum, zero weight decay and batch size 32.
The reason for zero weight decay is that the number of parameters in the proposed network is much less than that of VGG16. 
Thus, there is no need to punish.
The initial learning rate is 0.1.
We train 1,500,000 iterations and reduce the learning rate by multiplying 0.1 at iteration 600,000, 1,000,000, 1,200,000 and 1,400,000.
The training time is about 5 days with two NVIDIA GTX1080TI.
Our method is implemented using MXNet~\cite{MXNET} and the source code is released\footnote{https://github.com/YonghaoHe/A-Light-and-Fast-Face-Detector-for-Edge-Devices}.

\section{Experiments} 
In this section, comprehensive and extensive experiments are conducted.
Firstly, a new evaluation schema is proposed and the evaluation results on benchmarks are presented.
Secondly, we analyse the running efficiency on multiple platforms.
Thirdly, we further investigate the amount of computation and storage memory cost, introducing the computation efficiency rate.

\begin{figure}[ht]
\subfigure[Discontinuous ROC curves]{
\centering
\includegraphics[width=9cm]{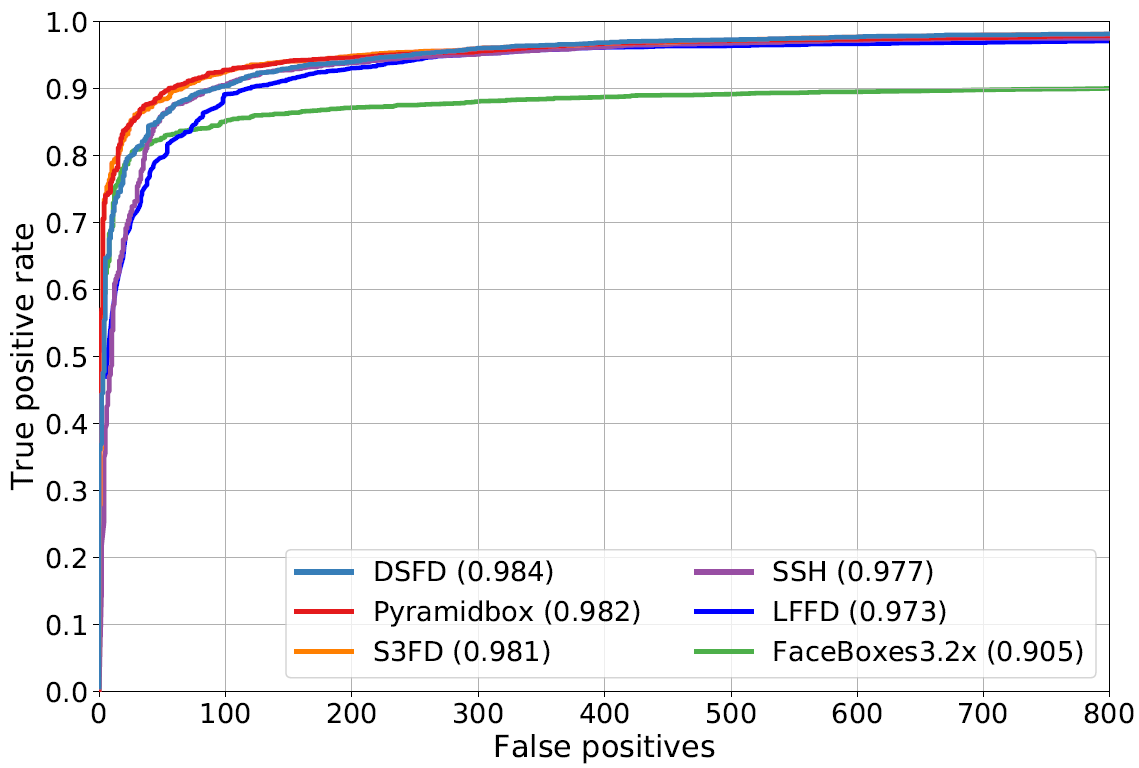}
}

\subfigure[Continuous ROC curves]{
\centering
\includegraphics[width=9cm]{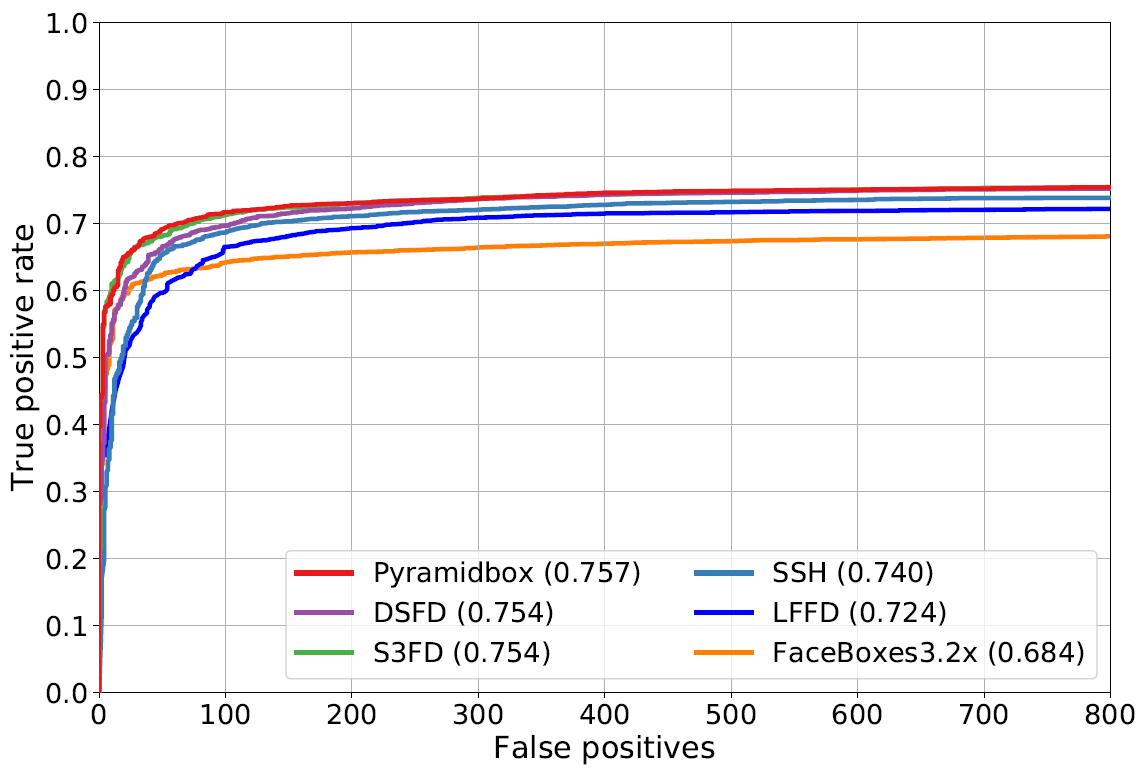}
}

\caption{Evaluation results on FDDB. Many other published methods are not displayed here for clarity.}
\label{fig:FDDB_results}
\end{figure}

\subsection{Evaluation on Benchmarks}
In this subsection, a new evaluation schema is described at the beginning.
The new schema is named as Single Inference on the Original (\textbf{SIO}).
SIO is proposed to reform the evaluation procedure for real-world applications.
We notice that the latency in some practical scenarios has the same importance as the accuracy.
The conventional evaluation procedure involves some tricky means, such as flips and image pyramids, for achieving higher accuracy.
However, the time consumption is not acceptable by doing that.
To this end, SIO can be easily operated in the following way: 
1) keep the image in its original size as the net input;
2) the net does only one inference with the original image.
The outputs of SIO are fed to the subsequent metrics.

In the experiments, we have to reproduce the results according to SIO schema. 
Therefore, we collect the compared methods which have released codes and models.
Finally, the following methods are taken for comparison:
DSFD~\cite{DSFD} ( Resnet152 backbone ), PyramidBox~\cite{PyramidBox} ( VGG16 backbone ), S3FD~\cite{S3FD} ( VGG16 backbone ), SSH~\cite{SSH} ( VGG16 backbone ) and FaceBoxes~\cite{FaceBoxes}.
DSFD and PyramidBox are state of the art methods.
The proposed method is named as LFFD.
LFFD and FaceBoxes do not rely on existing pre-trained backbones and are trained from scratch.
We evaluate all methods on two benchmarks: FDDB~\cite{FDDB} and WDIER FACE~\cite{WIDER}.

\textbf{FDDB dataset.}
FDDB contains 2845 images with 5171 unconstrained faces.
There are two types of scoring: discrete score and continuous score.
The first scoring criterion is obtained by thresholding IOU. 
And the second criterion directly uses IOU ratios.
We show final evaluation results of LFFD on FDDB against above five methods in Fig.~\ref{fig:FDDB_results}.
The overall performance on both scoring types shows the similar trends.
DSFD, PyramidBox, S3FD and SSH can achieve high accuracy with marginal gaps.
The proposed LFFD gains slightly lower accuracy than the first four methods, but outperforms FaceBoxes evidently.
The results indicate that LFFD is superior for detecting unconstrained faces. 

\begin{table}
\begin{center}
\begin{tabular}{c}
\includegraphics[width=8.2cm]{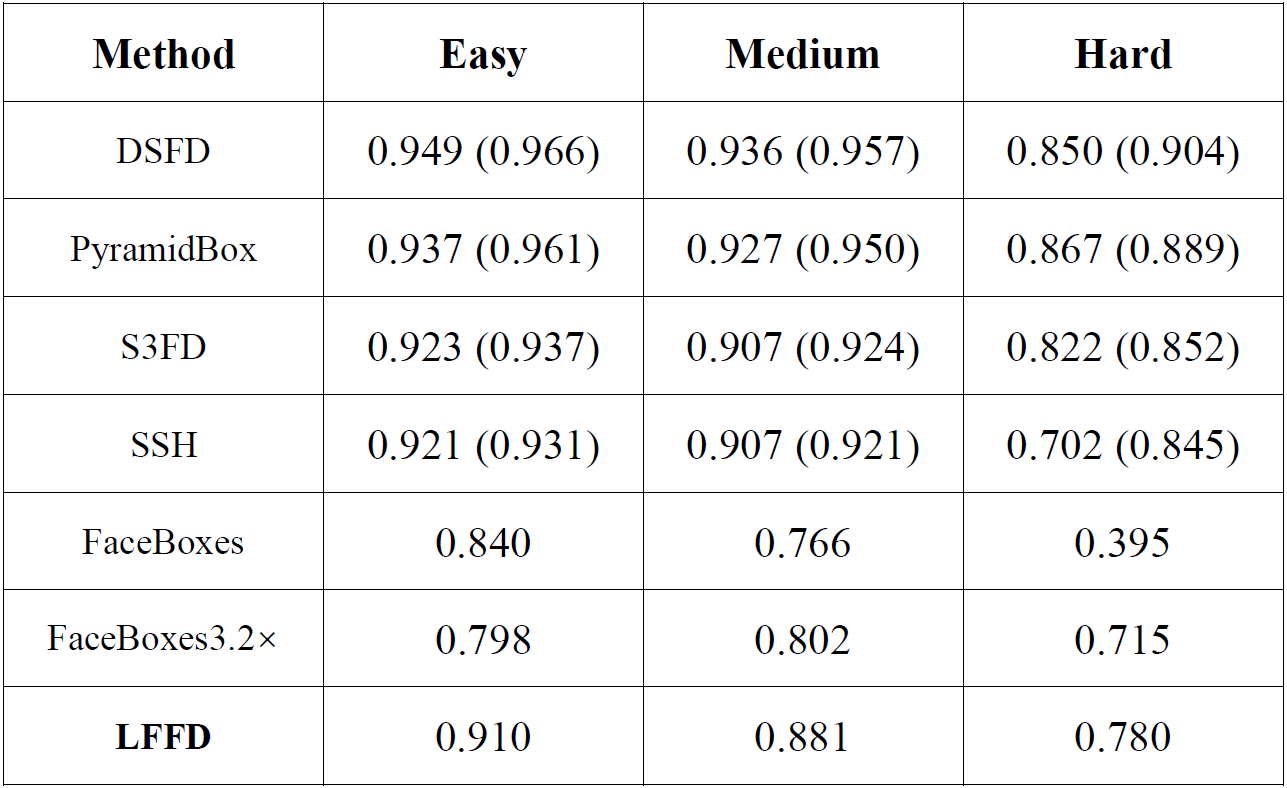}
\end{tabular}
\end{center}
\caption{Performance results on the \textbf{validation} set of WIDER FACE. The values in () are results from the original papers.}
\label{table_wider_val}
\end{table}

\begin{table}
\begin{center}
\begin{tabular}{c}
\includegraphics[width=8.2cm]{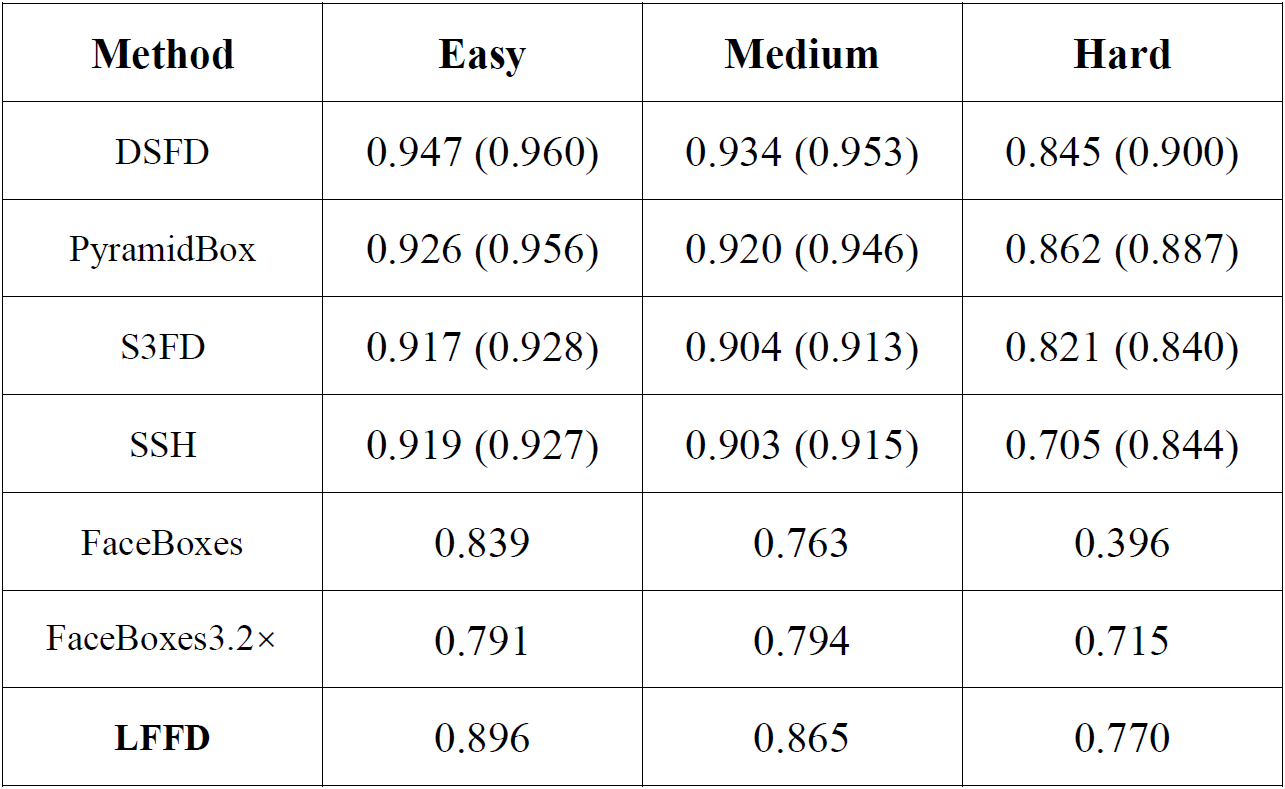}
\end{tabular}
\end{center}
\caption{Performance results on the \textbf{testing} set of WIDER FACE. The values in () are results from the original papers.}
\label{table_wider_test}
\end{table}

\textbf{WIDER FACE dataset.}
In WIDER FACE, there are 32,203 images and 393,703 labelled faces.
These faces are in a high degree of variability in scale, pose and occlusion.
Until now, WIDER FACE is the most widely used benchmark for face detection.
All images are randomly divided into three subsets: training set (40\%), validation set (10\%) and testing set(50\%).
Furthermore, images in each subset are graded to three levels (Easy, Medium and Hard) according to the difficulties for detection.
Roughly speaking, a large number of tiny/small faces are in Medium and Hard parts.
The groundtruth annotations are available only for training and validation sets.
All the compared methods are trained on training set.
We report the results on the validation and testing sets in Table~\ref{table_wider_val} and~\ref{table_wider_test}, respectively.

Some observations can be made. 
Firstly, performance drop is evident for DSFD, PyramidBox, S3FD and SSH compared to their original results.
On the one hand, achieving high accuracy through only one inference is relatively difficult.
On the other hand, the tricks can indeed improve the accuracy impressively.
Secondly, PyramidBox obtains the best results on Hard parts,
whereas the performance of SSH on Hard parts is decreased dramatically mainly due to the neglect of some tiny faces.
Thirdly, FaceBoxes does not get desirable results on Medium and Hard parts.
Since FaceBoxes produces large stride 32 rapidly, which means that faces smaller than 32 pixels are hardly detected.
To make it clearer, we conduct additional experiments for FaceBoxes, named as FaceBoxes3.2$\times$, in which the both sides of input images are enlarged 3.2$\times$.
We can see that the results on Medium and Hard parts are improved remarkably.
The performance drop on Easy parts is attributed to that some faces are resized too large to be detected.
To some extent, the results of FaceBoxes and FaceBoxes3.2$\times$ indicate that FaceBoxes can not cover faces with large range.
Fourthly, the proposed method LFFD consistently outperforms FaceBoxes, although having gaps with state of the art methods.
Additionally, LFFD is better than SSH that uses VGG16 as the backbone on Hard parts.

\subsection{Running Efficiency}
In this subsection, we analyse the running speed of all methods on three different platforms.
The information of each platform and related libraries are listed in Table~\ref{table_hardware_platform}.
We use batchsize 1 and a few common resolutions for testing.
For fair comparison, FaceBoxes3.2$\times$ is used here instead of FaceBoxes.
The running speed is measured in \textit{ms} and the corresponding \textit{FPS}.
The final results are presented in Table~\ref{table_titanxp_running_efficiency},~\ref{table_tx2_running_efficiency} and~\ref{table_arm_running_efficiency}.

In Table~\ref{table_titanxp_running_efficiency}, we also add VGG16 and Resnet50 for sufficient comparison.
SSH and S3FD are based on VGG16, having similar speed with VGG16.
Whereas, PyramidBox is much slower due to additional complex modules, although based on VGG16 as well.
DSFD can achieve state of the art accuracy, but it has the slowest running speed.
The proposed LFFD runs the fastest at 3840$\times$2160, and FaceBoxes3.2$\times$ obtains the highest speed at other three resolutions.
Both LFFD and FaceBoxes3.2$\times$ can reach or even exceed the real-time running speed ($>$ 30 FPS) at the first three resolutions.
The aforementioned trend that state of the art methods pursue higher accuracy at the cost of running speed is clearly verified.

TX2 and Raspberry Pi 3 are edge devices with low computation power.
DSFD, PyramidBox, S3FD and SSH are either too slow or failed to run on these two platforms.
Thus, we only evaluate the proposed LFFD and FaceBoxes3.2$\times$ at lower resolutions in Table~\ref{table_tx2_running_efficiency} and~\ref{table_arm_running_efficiency}.
The overall results show that LFFD is faster than FaceBoxes3.2$\times$ except for the case at 640$\times$480 on Raspberry Pi 3.
LFFD can better benefit from optimizations of ncnn than FaceBoxes3.2$\times$ at low resolutions 160$\times$120 and 320$\times$240.

\begin{table}
\begin{center}
\begin{tabular}{c}
\includegraphics[width=8.2cm]{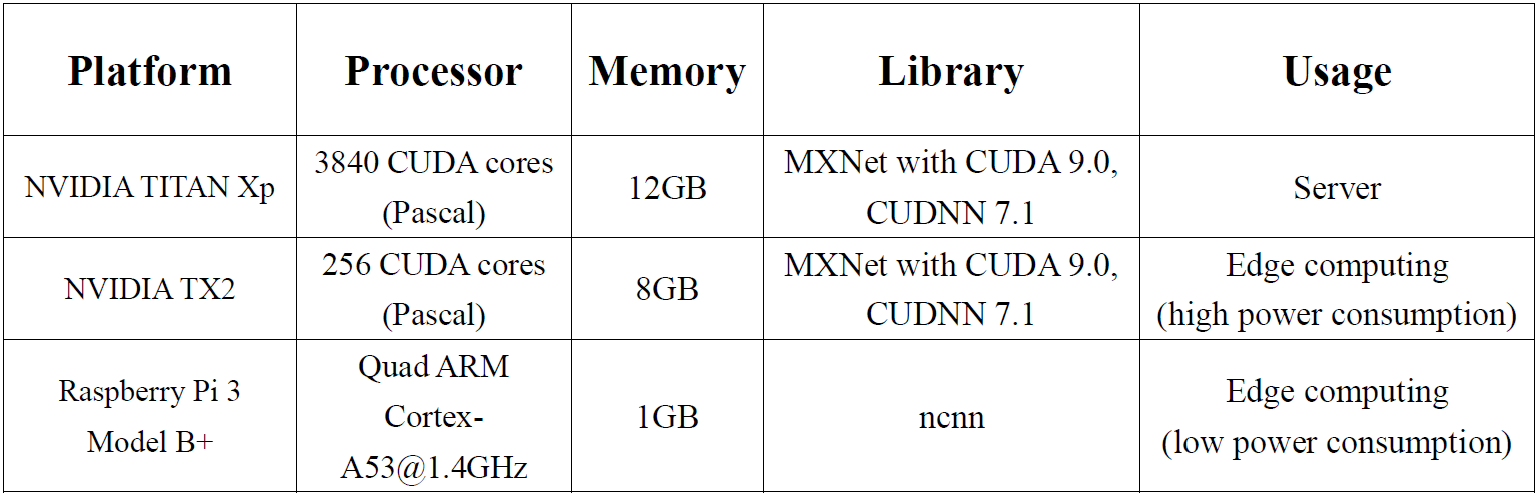}
\end{tabular}
\end{center}
\caption{Information of hardware platforms and related running libraries.}
\label{table_hardware_platform}
\end{table}

\begin{table}
\begin{center}
\begin{tabular}{c}
\includegraphics[width=8.2cm]{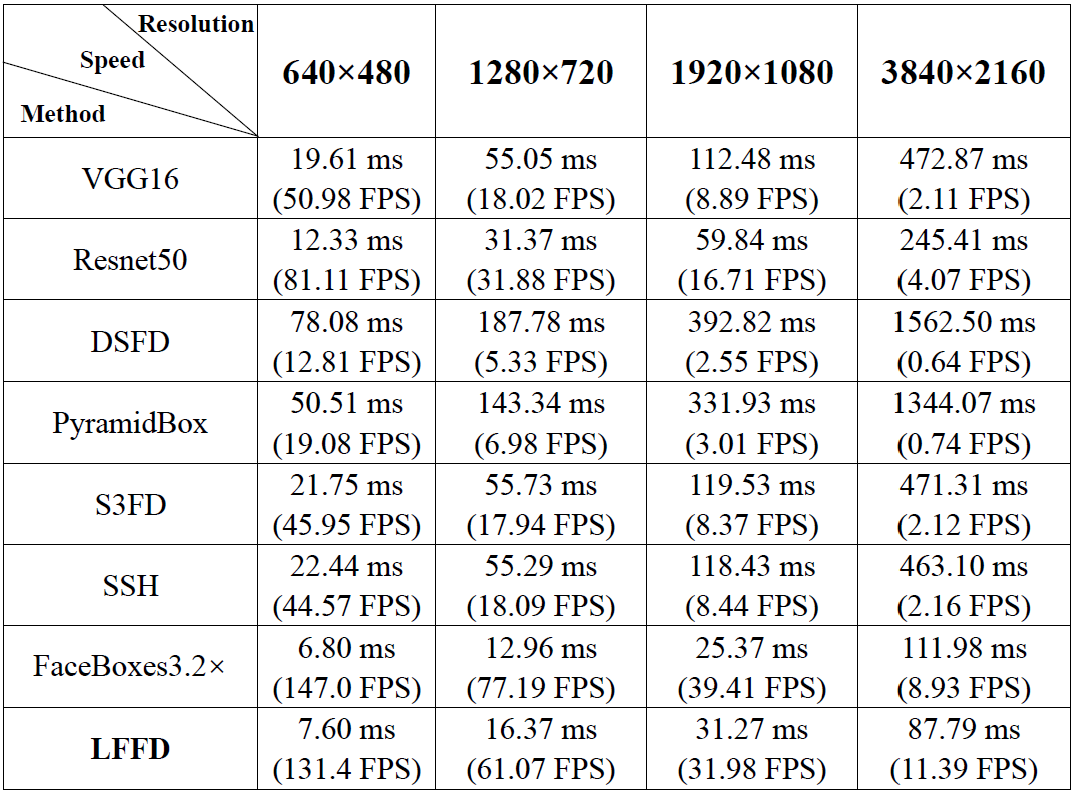}
\end{tabular}
\end{center}
\caption{Running efficiency on TITAN Xp.}
\label{table_titanxp_running_efficiency}
\end{table}

\begin{table}
\begin{center}
\begin{tabular}{c}
\includegraphics[width=8.2cm]{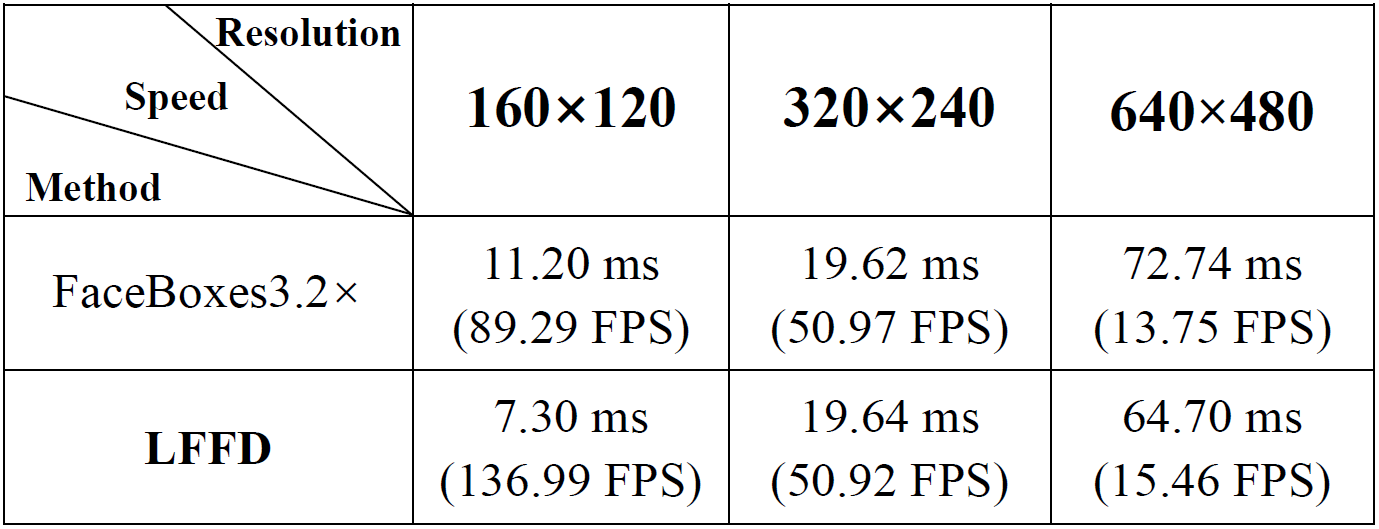}
\end{tabular}
\end{center}
\caption{Running efficiency on TX2.}
\label{table_tx2_running_efficiency}
\end{table}

\begin{table}
\begin{center}
\begin{tabular}{c}
\includegraphics[width=8.2cm]{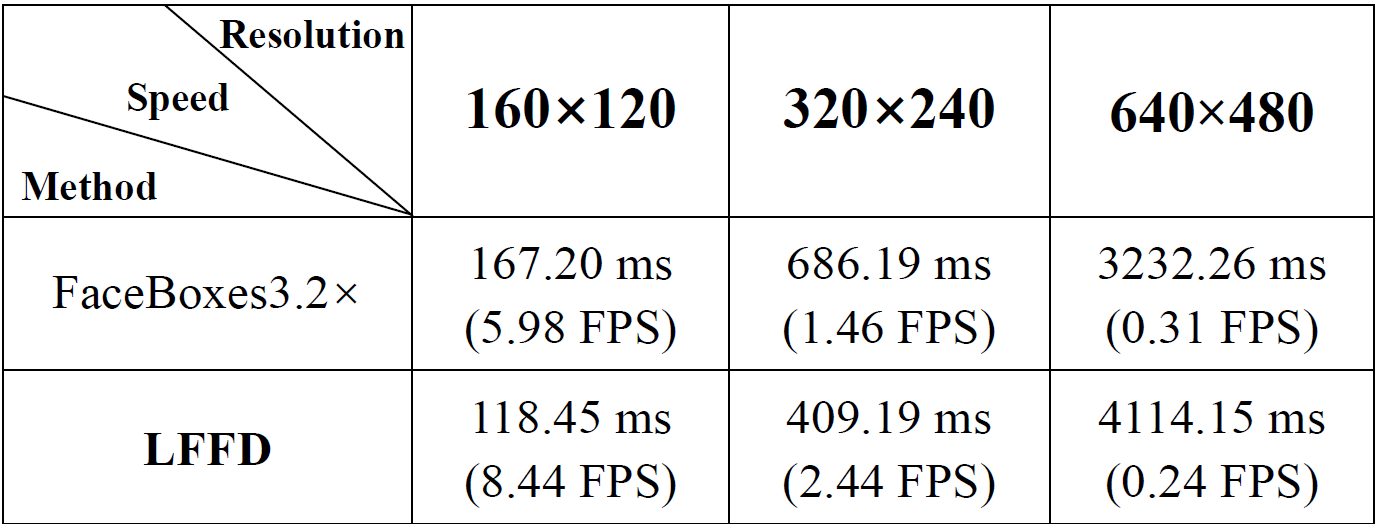}
\end{tabular}
\end{center}
\caption{Running efficiency on Raspberry Pi 3 Model B+.}
\label{table_arm_running_efficiency}
\end{table}

\subsection{Parameter, Computation and Model Size}
We investigate the compared methods from the perspective of parameter, computation and model size in this subsection.
The edge devices always have constrained storage memories.
It is necessary to consider the memory usage of face detectors.
The number of parameters is highly related to the model size.
However, less parameters do not mean less computation.
Following~\cite{PCNNREI}, we use FLOPs to measure the computation at resolution 640$\times$480.
All the information is presented in Table~\ref{table_parameter_computation_modelsize}.

For state of the art methods DSFD and PyramidBox, they have large amounts of parameters and FLOPs.
The proposed LFFD and FaceBoxes3.2$\times$ have light networks which are appropriate to deploy on edge devices.
To further demonstrate the efficiency of the proposed network, we define a new metric:
\begin{eqnarray}
E_{net} = FLOPs/t,
\end{eqnarray}
where $t$ indicates the running time.
$E_{net}$ reflects the computation efficiency of networks (the larger, the more efficient) and can be calculated at a certain resolution on a specific platform.
We compute this metric for LFFD and FaceBoxes3.2$\times$ at 640$\times$480 on three platforms (LFFD vs. FaceBoxes3.2$\times$):
\begin{itemize}
\setlength{\itemsep}{0pt}
\setlength{\parsep}{0pt}
\setlength{\parskip}{0pt}
\item 1.22G/ms vs. 0.42G/ms on TITAN Xp; 
\item 0.14G/ms vs. 0.04G/ms on TX2; 
\item 0.0022G/ms vs. 0.00088G/ms on Raspberry                                                                                                                                                                                                                                                                                                                                                                                                                                                                                                                                                                                                                                     Pi 3; 
\end{itemize}
Evidently, the proposed network has much more efficient computation, which demonstrates the superiority of the concise network design.

\begin{table}
\begin{center}
\begin{tabular}{c}
\includegraphics[width=8.2cm]{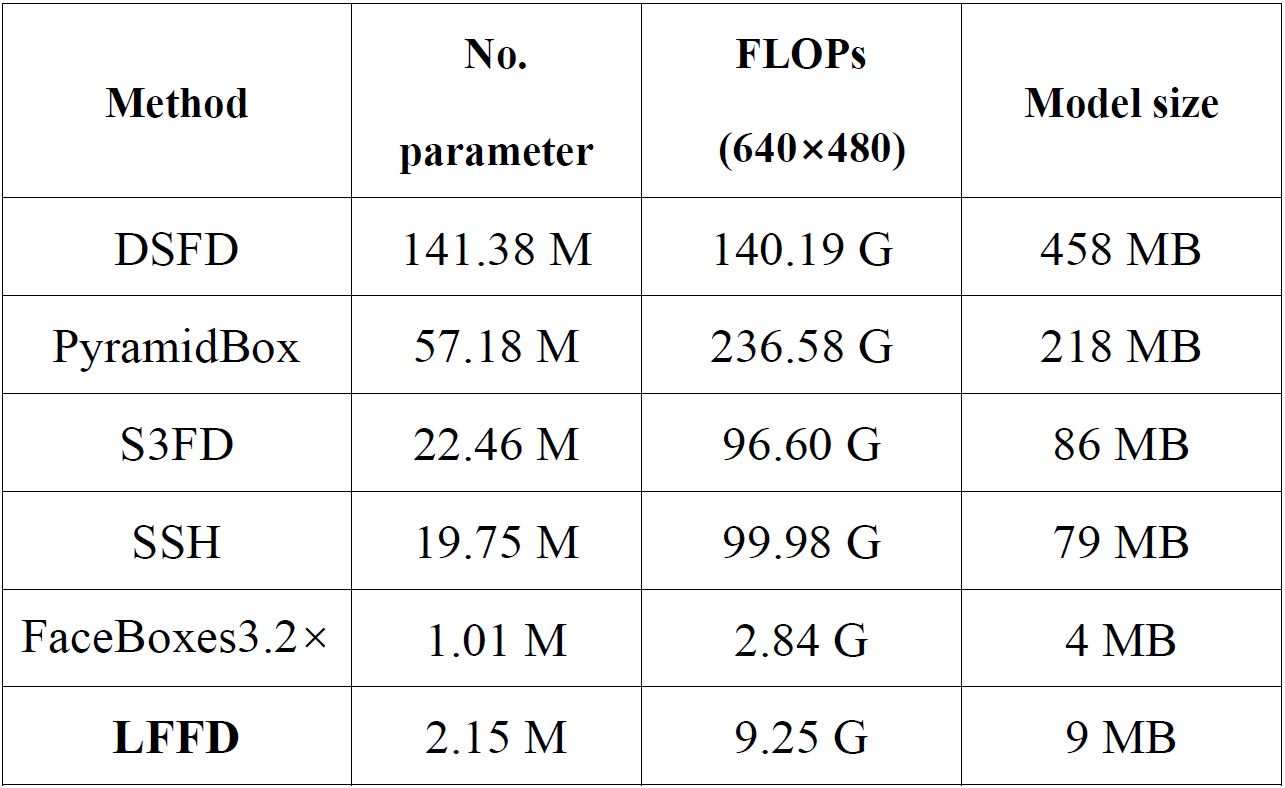}
\end{tabular}
\end{center}
\caption{Number of parameters, FLOPs and model size. The model size may vary slightly with different libraries.}
\label{table_parameter_computation_modelsize}
\end{table}

\section{Conclusion}
This paper introduces a light and fast face detector that properly balances accuracy and latency.
By deeply rethinking the RF in the background of face detection, we propose an anchor-free method to overcome the drawbacks of anchor-based methods. The proposed method regards the RFs as natural ``anchors'' which can cover continuous face scales and reach nearly 100\% hit rate.
After investigating the essential relations between ERFs and face scales, we delicately design an simple but efficient network with eight detecting branches.
The proposed network consists of common building blocks with less filters, resulting in fast inference speed.
Comprehensive and extensive experiments are conducted to fully analyse the proposed method.
The final results demonstrate that our method can achieve superior accuracy with small model size and efficient computation, which makes it an excellent candidate for edge devices.

{\small
\bibliographystyle{ieee}
\bibliography{egbib}

\begin{thebibliography}{10}\itemsep=-1pt

\bibitem{CBE}
S.~C. Brubaker, J.~Wu, J.~Sun, M.~D. Mullin, and J.~M. Rehg.
\newblock On the design of cascades of boosted ensembles for face detection.
\newblock {\em International Journal of Computer Vision}, 77:65--86, 2008.

\bibitem{MXNET}
T.~Chen, M.~Li, Y.~Li, M.~Lin, N.~Wang, M.~Wang, T.~Xiao, B.~Xu, C.~Zhang, and
  Z.~Zhang.
\newblock Mxnet: A flexible and efficient machine learning library for
  heterogeneous distributed systems.
\newblock {\em arXiv:1512.01274}, 2015.

\bibitem{SRN}
C.~Chi, S.~Zhang, J.~Xing, Z.~Lei, S.~Z. Li, and X.~Zou.
\newblock Selective refinement network for high performance face detection.
\newblock {\em arXiv:1809.02693}, 2018.

\bibitem{ACF}
P.~Dollár, R.~Appel, S.~Belongie, and P.~Perona.
\newblock Fast feature pyramids for object detection.
\newblock {\em IEEE Transactions on Pattern Analysis and Machine Intelligence},
  36(8):1532--1545, 2014.

\bibitem{FastRCNN}
R.~Girshick.
\newblock Fast r-cnn.
\newblock In {\em Proceedings of IEEE International Conference on Computer
  Vision}, pages 1440--1448, 2015.

\bibitem{RCNN}
R.~Girshick, J.~Donahue, T.~Darrell, and J.~Malik.
\newblock Rich feature hierarchies for accurate object detection and semantic
  segmentation.
\newblock In {\em Proceedings of IEEE Conference on Computer Vision and Pattern
  Recognition}, pages 580--587, 2014.

\bibitem{Resnet}
K.~He, X.~Zhang, S.~Ren, and J.~Sun.
\newblock Deep residual learning for image recognition.
\newblock In {\em Proceedings of IEEE Conference on Computer Vision and Pattern
  Recognition}, pages 770--778, 2016.

\bibitem{SIDCNN}
A.~G. Howard.
\newblock Some improvements on deep convolutional neural network based image
  classification.
\newblock {\em arXiv:1312.5402}, 2013.

\bibitem{HR}
P.~Hu and D.~Ramanan.
\newblock Finding tiny faces.
\newblock In {\em Proceedings of IEEE Conference on Computer Vision and Pattern
  Recognition}, pages 951--959, 2017.

\bibitem{Densenet}
G.~Huang, Z.~Liu, L.~van~der Maaten, and K.~Q. Weinberger.
\newblock Densely connected convolutional networks.
\newblock In {\em Proceedings of IEEE Conference on Computer Vision and Pattern
  Recognition}, pages 4700--4708, 2017.

\bibitem{BN}
S.~Ioffe and C.~Szegedy.
\newblock Batch normalization: Accelerating deep network training by reducing
  internal covariate shift.
\newblock {\em arXiv:1502.03167}, 2015.

\bibitem{FDDB}
V.~Jain and E.~Learned-Miller.
\newblock Fddb: A benchmark for face detection in unconstrained settings.
\newblock Technical report, University of Massachusetts, Amherst, 2010.

\bibitem{FFRCNN}
H.~Jiang and E.~Learned-Miller.
\newblock Face detection with the faster r-cnn.
\newblock In {\em Proceedings of IEEE International Conference on Automatic
  Face \& Gesture Recognition}, pages 650--657, 2017.

\bibitem{FAS}
X.~Jin and X.~Tan.
\newblock Face alignment in-the-wild: A survey.
\newblock {\em Computer Vision and Image Understanding}, 162:1--22, 2017.

\bibitem{ICDCNN}
A.~Krizhevsky, I.~Sutskever, and G.~E. Hinton.
\newblock Imagenet classification with deep convolutional neural networks.
\newblock In {\em Proceedings of Advances in Neural Information Processing
  Systems}, pages 1097--1105, 2012.

\bibitem{DSFD}
J.~Li, Y.~Wang, C.~Wang, Y.~Tai, J.~Qian, J.~Yang, C.~Wang, J.~Li, and
  F.~Huang.
\newblock Dsfd: dual shot face detector.
\newblock In {\em Proceedings of IEEE Conference on Computer Vision and Pattern
  Recognition}, 2019.

\bibitem{SSD}
W.~Liu, D.~Anguelov, D.~Erhan, C.~Szegedy, S.~Reed, C.-Y. Fu, and A.~C. Berg.
\newblock Ssd: Single shot multibox detector.
\newblock In {\em Proceedings of European Conference on Computer Vision}, pages
  21--37, 2016.

\bibitem{UERF}
W.~Luo, Y.~Li, R.~Urtasun, and R.~Zemel.
\newblock Understanding the effective receptive field in deep convolutional
  neural networks.
\newblock In {\em Proceedings of Advances in Neural Information Processing
  Systems}, pages 4898--4906, 2016.

\bibitem{PCNNREI}
P.~Molchanov, S.~Tyree, T.~Karras, T.~Aila, and J.~Kautz.
\newblock Pruning convolutional neural networks for resource efficient
  inference.
\newblock {\em arXiv:1611.06440}, 2016.

\bibitem{SSH}
M.~Najibi, P.~Samangouei, R.~Chellappa, and L.~S. Davis.
\newblock Ssh: Single stage headless face detector.
\newblock In {\em Proceedings of IEEE International Conference on Computer
  Vision}, pages 4875--4884, 2017.

\bibitem{LBP}
T.~Ojala, M.~Pietikäinen, and T.~Mäenpää.
\newblock Multiresolution gray-scale and rotation invariant texture
  classification with local binary patterns.
\newblock {\em IEEE Transactions on Pattern Analysis and Machine Intelligence},
  24:971--987, 2002.

\bibitem{SBFD}
M.-T. Pham and T.-J. Cham.
\newblock Fast training and selection of haar features using statistics in
  boosting-based face detection.
\newblock In {\em Proceedings of IEEE International Conference on Computer
  Vision}, pages 1--7, 2007.

\bibitem{YOLO}
J.~Redmon, S.~Divvala, R.~Girshick, and A.~Farhadi.
\newblock You only look once: Unified, real-time object detection.
\newblock In {\em Proceedings of IEEE Conference on Computer Vision and Pattern
  Recognition}, pages 779--788, 2016.

\bibitem{YOLOv2}
J.~Redmon and A.~Farhadi.
\newblock Yolo9000: better, faster, stronger.
\newblock In {\em Proceedings of IEEE Conference on Computer Vision and Pattern
  Recognition}, pages 7263--7271, 2017.

\bibitem{YOLOv3}
J.~Redmon and A.~Farhadi.
\newblock Yolov3: An incremental improvement.
\newblock {\em arXiv:1804.02767}, 2018.

\bibitem{FasterRCNN}
S.~Ren, K.~He, R.~Girshick, and J.~Sun.
\newblock Faster r-cnn: Towards real-time object detection with region proposal
  networks.
\newblock In {\em Proceedings of Advances in Neural Information Processing
  Systems}, pages 91--99, 2015.

\bibitem{VGG}
K.~Simonyan and A.~Zisserman.
\newblock Very deep convolutional networks for large-scale image recognition.
\newblock {\em arXiv:1409.1556}, 2014.

\bibitem{PyramidBox}
X.~Tang, D.~K. Du, Z.~He, and J.~Liu.
\newblock Pyramidbox: A context-assisted single shot face detector.
\newblock In {\em Proceedings of European Conference on Computer Vision}, pages
  797--813, 2018.

\bibitem{Viola-Jones}
P.~Viola and M.~J. Jones.
\newblock Robust real-time face detection.
\newblock {\em International Journal of Computer Vision}, 57(2):137--154, 2004.

\bibitem{FRCNN}
H.~Wang, Z.~Li, X.~Ji, and Y.~Wang.
\newblock Face r-cnn.
\newblock {\em arXiv:1706.01061}, 2017.

\bibitem{DFRS}
M.~Wang and W.~Deng.
\newblock Deep face recognition: A survey.
\newblock {\em arXiv:1804.06655}, 2018.

\bibitem{ACF-Face}
B.~Yang, J.~Yan, Z.~Lei, and S.~Z. Li.
\newblock Aggregate channel features for multi-view face detection.
\newblock In {\em Proceedings of IEEE International Joint Conference on
  Biometrics}, pages 1--8, 2014.

\bibitem{WIDER}
S.~Yang, P.~Luo, C.~C. Loy, and X.~Tang.
\newblock Wider face: A face detection benchmark.
\newblock In {\em Proceedings of IEEE Conference on Computer Vision and Pattern
  Recognition}, pages 5525--5533, 2016.

\bibitem{MTCNN}
K.~Zhang, Z.~Zhang, Z.~Li, and Y.~Qiao.
\newblock Joint face detection and alignment using multitask cascaded
  convolutional networks.
\newblock {\em IEEE Signal Processing Letters}, 23(10):1499--1503, 2016.

\bibitem{ICS}
K.~Zhang, Z.~Zhang, H.~Wang, Z.~Li, Y.~Qiao, and W.~Liu.
\newblock Detecting faces using inside cascaded contextual cnn.
\newblock In {\em Proceedings of IEEE International Conference on Computer
  Vision}, pages 3171--3179, 2017.

\bibitem{ISRN}
S.~Zhang, R.~Zhu, X.~Wang, H.~Shi, T.~Fu, S.~Wang, T.~Mei, and S.~Z. Li.
\newblock Improved selective refinement network for face detection.
\newblock {\em arXiv:1901.06651}, 2019.

\bibitem{FaceBoxes}
S.~Zhang, X.~Zhu, Z.~Lei, H.~Shi, X.~Wang, and S.~Z. Li.
\newblock Faceboxes: A cpu real-time face detector with high accuracy.
\newblock In {\em Proceedings of IEEE International Joint Conference on
  Biometrics}, pages 1--9, 2017.

\bibitem{S3FD}
S.~Zhang, X.~Zhu, Z.~Lei, H.~Shi, X.~Wang, and S.~Z. Li.
\newblock S3fd: Single shot scale-invariant face detector.
\newblock In {\em Proceedings of IEEE International Conference on Computer
  Vision}, pages 192--201, 2017.

\bibitem{VIM-FD}
Y.~Zhang, X.~Xu, and X.~Liu.
\newblock Robust and high performance face detector.
\newblock {\em arXiv:1901.02350}, 2019.

\bibitem{SSF}
C.~Zhu, R.~Tao, K.~Luu, and M.~Savvides.
\newblock Seeing small faces from robust anchor's perspective.
\newblock In {\em Proceedings of IEEE Conference on Computer Vision and Pattern
  Recognition}, pages 5127--5136, 2018.

\bibitem{HOG}
Q.~Zhu, M.-C. Yeh, K.-T. Cheng, and S.~Avidan.
\newblock Fast human detection using a cascade of histograms of oriented
  gradients.
\newblock In {\em Proceedings of IEEE Conference on Computer Vision and Pattern
  Recognition}, pages 1491--1498, 2006.

\end{thebibliography}
}

\end{document}